# BINARY SINE COSINE ALGORITHMS FOR FEATURE SELECTION FROM MEDICAL DATA

Shokooh Taghian[1,2] and Mohammad H. Nadimi-Shahraki[1,2,*]

[1]Faculty of Computer Engineering, Najafabad Branch, Islamic Azad University, Najafabad, Iran
[2]Big Data Research Center, Najafabad Branch, Islamic Azad University, Najafabad, Iran

## ABSTRACT

*A well-constructed classification model highly depends on input feature subsets from a dataset, which may contain redundant, irrelevant, or noisy features. This challenge can be worse while dealing with medical datasets. The main aim of feature selection as a pre-processing task is to eliminate these features and select the most effective ones. In the literature, metaheuristic algorithms show a successful performance to find optimal feature subsets. In this paper, two binary metaheuristic algorithms named S-shaped binary Sine Cosine Algorithm (SBSCA) and V-shaped binary Sine Cosine Algorithm (VBSCA) are proposed for feature selection from the medical data. In these algorithms, the search space remains continuous, while a binary position vector is generated by two transfer functions S-shaped and V-shaped for each solution. The proposed algorithms are compared with four latest binary optimization algorithms over five medical datasets from the UCI repository. The experimental results confirm that using both bSCA variants enhance the accuracy of classification on these medical datasets compared to four other algorithms.*

## 1. INTRODUCTION

By advancing in the technology, a massive amount of data is regularly generated and stored from real-world applications such as medical, transportation, tourism and engineering. This massive data contains a large number of different features. However, not all the features are needed for analyzing and discovering knowledge, since many of them are redundant or irrelevant to the problem. Many redundant or irrelevant features are not effective for solving classification problems; moreover, they may increase the computational complexity and decrease the classification accuracy [1, 2].

Dimensionality reduction is one of the most important preprocessing techniques, which aims to reduce the number of features under some criterion and obtain a better performance. One of the most important tools in dimension reduction is the feature selection [3]. Feature selection is the process of selecting more effective and relevant features in order to reduce the dimensionality of data and improve the classification performance [4]. As shown in Fig. 1, feature selection has four main phases including, subset generation, subset evaluation, stopping criteria, and validation [5]. In the first step, the search strategy employs different methods in order to generate a new subset as a solution. The second step includes a classifier and a predefined fitness function to evaluate the quality of the generated solutions. This process continues until the termination criteria are met. In the literature, there are two different approaches filter and wrapper to select the effective features [6]. The former approach uses measures such as distance, dependency, or consistency of the features to find the optimal subset. The latter uses a specific classifier to evaluate the quality of selected features and find the near-optimal solutions from an exponential set of features. The major drawback of filter method is that it lacks the influence of features on the performance of the classifier [7]; however, since it does not use the learning algorithms, it is usually fast and suitable for use with large data sets. On the other hand, the wrapper method is known to be more accurate but it is computationally more expensive [8].

The feature selection is to find the optimum combination of features; therefore, it can be considered as a search process. Since evaluating $2^N$ -1 subsets of a dataset with N features is an NP-hard problem, finding the best subset cannot be achieved using an exhaustive search algorithm. Metaheuristic algorithms are known for their ability in finding near-optimum solutions for global optimization problems within a reasonable time. These algorithms can exploit the solution with good fitness and have the potential of finding promising areas.

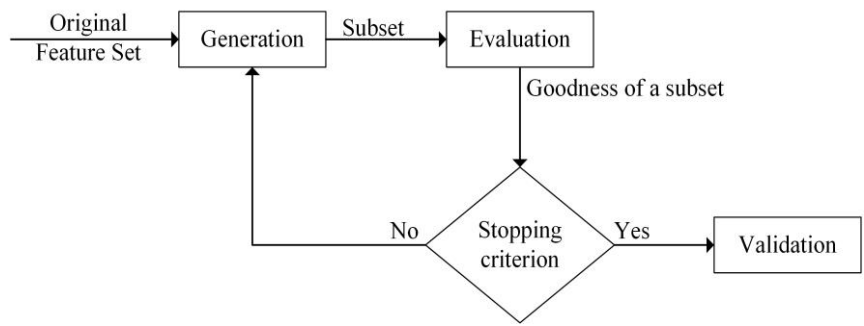

Figure 1. Feature selection process

However, due to the random nature of the metaheuristic algorithms, there is no guarantee that they can find the optimal feature subset for different problems [9]. Additionally, according to the No-Free-Lunch theorem [10], there is no single, all-purpose, and general optimization algorithm, which can find optimum solutions for all problems. Therefore, many metaheuristic algorithms have been proposed for solving continuous problems such as particle swarm optimization (PSO) [11], differential evolution (DE) [12], artificial bee colony (ABC) [13], bat algorithm (BA) [14], gravitational search algorithm (GSA) [15], grey wolf optimizer (GWO) [16], and sine cosine algorithm (SCA) [17]. Also, with increasing the number of variables and complexity of the problems, the high dimensional problems are an emerging issue, and recently some metaheuristic algorithms such as conscious neighborhood-based crow search algorithm (CCSA) [18] have been proposed for solving large-scale optimization problems. Some algorithms such as genetic algorithm [19] and ant colony algorithm (ACO) [20] were proposed for solving the discrete optimization problems. Meanwhile, different methods were introduced to adapt a continuous metaheuristic algorithm for a discrete search space [21]. Because of having the successful results of metaheuristic algorithms, they are widely applied to solve a variety of discrete and continuous optimization [22-26].

Sine Cosine Algorithm was recently proposed for continuous optimization problems which attracts the attention of many researchers to use its potentials and apply to different applications. Although some binary variants of the SCA were proposed for discrete optimization problems, there is no variant of this algorithm for feature selection from medical datasets. This is our motivation to develop another binary version of the SCA.

The rest of this paper is organized as follow: a review of the literature on binary metaheuristic algorithms used in feature selection problem is explained in Section 2. In Section 3 the continuous Sine Cosine Algorithm is described. The proposed binary versions of sine cosine algorithm describe in Section4. The proposed algorithms for feature selection problem are presented in Section 5 and the experimental results are reported in Section 6. Finally, the conclusion and future works are stated in Section 7 contains.

## 2. RELATED WORKS

In the past decade, metaheuristic algorithms attract attention of many researchers due to their powerful and efficient performance in dealing with complex real-world problems [27]. A great deal of efforts has been made to solve various problems in different fields. However, many well-known metaheuristic algorithms are designed for solving continuous problems, while some problems have a binary nature. Therefore, binary versions of these algorithms were developed to solve these problems. Most of the well-known metaheuristic algorithms have the binary version which makes them capable of solving binary problems.

In the literature, different methods exist to develop a binary algorithm such as normalization, rounding, considering a binary search space, and using binary operators. In addition, the transfer function is another method that is used to modify the value of continuous components into binary values [21]. BGSA [28] is a binary version of the GSA, which used a V-shaped transfer function applied on the velocity parameter in order to calculate the mass movement probability. In [29], a binary version of GWO was combined with KNN classifier to calculate the fitness function of each features subset. An S-shaped transfer function is applied on the position of each wolf to estimate the position changing. In [30], the continuous Dragonfly algorithm (DA) [31] was modified to tackle the feature selection problem. This is performed by using the V-shaped transfer function that is applied on the step vector value of each search agent. Binary Salp Swarm Algorithm (SSA) [32] is a recent binary metaheuristic algorithm, that uses S-shaped and V-shaped transfer functions to modify the algorithm in order to solve feature selection problems. Lately, Whale Optimization Algorithm [33] has been employed as a feature selection algorithm for disease detection [34]. The binary butterfly [35], hybrid GWO with CSA [36], and evolutionary GSA [37] are some example of newly proposed binary algorithms that used for feature selection problem. For the SCA, two other variants are proposed with binary variables. In [38], a binary version of SCA is proposed by using the rounding method. Variables of this algorithm are bounded to 0 and 1, therefore each value of the solution is rounded to the nearest value to show the feature is selected or not. In the other work [39], a modified sigmoid function used as a mapped function to solve binary problems.

In this work, the focus is on using transfer functions to produce a binary version of the SCA for wrapper feature selection. The proposed algorithms select the optimal subset of features which increase the accuracy of the classifier and at the same time decrease the length of feature subset. The application of proposed algorithms then applied for disease detection by using UCI medical datasets [40].

## 3. THE SINE COSINE ALGORITHM (SCA)

The SCA is a population-based metaheuristic algorithm introduced for solving continuous optimization problems. The SCA starts with randomly distributing the solutions in the search space. After calculating the fitness value of each solution, the solution with the best fitness is considered as a destination solution. The destination solution is used in a position update equation shown by Eq. 1 by which the position of other solutions is changed.

$$X_i^{t+1} = \begin{cases} X_i^t + r_1 \times \sin(r_2) \times \left| r_3 P_i^t - X_i^t \right| & ,r_4 < 0.5 \\ X_i^t + r_1 \times \cos(r_2) \times \left| r_3 P_i^t - X_i^t \right| & ,r_4 \geq 0.5 \end{cases} \qquad (1)$$

where $X_i^t$ is the value of i-th dimension of the current solution at iteration t, $P_i^t$ is the position of the destination solution in i-th dimension and t-th iteration, and $r_1$, $r_2$, and $r_3$ are random numbers. In this equations, one formula is selected by a random number $r_4$, which is uniformly distributed between 0 and 1. The SCA runs until the termination criteria is met.

The SCA controls the exploration and exploitation of the algorithm and direct the solution to the next position using three parameters $r_1$, $r_2$, and $r_3$. The parameter $r_1$ has the ability of balancing the exploration and exploitation in the early and last stages of the SCA. This parameter determines the direction of the new solution either toward or outward the destination solution. It directs the search process, whether to explore the entire search space even far from the destination solution in the early stages of the algorithm or to exploit near the destination solution in order to find better solutions in the last stages of the algorithm. If $r_1 < 0$ the distance between the solution and the destination solution will be decreased, while it will be increased if $r_1 > 0$. The $r_1$ parameter is calculated by considering the maximum iterations T, the current iteration t, and a constant value a as shown in Eq. 4.

$$r_1 = a - t \frac{a}{t} \qquad (2)$$

The random parameter $r_2$ indicates the distance value of the solution from the destination solution position. The higher value of this parameter leads to exploration because the distance between the solution and the destination solution is more, while the lower value indicates the less distance and leads to exploitation. The third parameter $r_3$ is a weight to show the impact of the destination solution in defining the distance.

## 4. BINARY SINE COSINE ALGORITHM (BSCA)

The original SCA is to solve the continuous optimization problems where each individual can move freely in the entire search space; while a binary search space can be assumed as a hypercube that the individuals can only move to neither nearer or farther corners of the hypercube by flipping the bit-string position value [41]. Therefore, to use the SCA for solving binary problems, it must map the continuous values into the probability values using a transfer function in order to determine the binary position values. As discussed in our previous work [our work], two introduced families of the transfer functions are S-shaped and V-shaped. In this work, the transfer functions are utilized to convert the continuous SCA to binary versions which named SBSCA and VBSCA.

### 4.1. S-shaped Binary Sine Cosine Algorithm (SBSCA)

The search space in the proposed algorithms is considered as a continuous space in which each individual has a floating-point position vector. Therefore, to generate the individual's binary bit-string, the continuous values must be converted. The conversion is applied by using an S-shaped transfer function on each dimension of the position to force the individual to move in a binary space. The transfer function uses the floating-point position values to determine a bounded probability in the interval of [0, 1] for each individual. The probabilities then are used to generate bit-string position vector from a floating-point vector. The equation and the shape of the S-shaped function are given in Eq. 3 and Fig 2a.

$$S(x_i^d(t+1)) = \frac{1}{1+e^{-x_i^d(t)}} \tag{3}$$

The value of $S(x_i^d(t+1))$ indicates the probability of changing the binary position value of i-th individual in d-th dimension. Then, the probability, as mentioned in Eq. 4 compared with a threshold value to determine the binary value.

$$x_i^d(t+1) = \begin{cases} 1 & ,if\ rand \le S(x_i^d(t+1)) \\ 0 & ,otherwise \end{cases} \tag{4}$$

## 4.2. V-shaped Binary Sine Cosine Algorithm (VBSCA)

The V-shaped transfer function is the other function which is used for calculating the position changing probabilities. The V-shaped transfer function shown in Fig. 2b, like the S-shaped transfer function, is first utilized for calculating the probability of changing the individual's positions by Eq.5.

$$V(x_i^d(t+1)) = \left| \frac{2}{\pi}\arctan(\frac{\pi}{2})(x_i^d(t)) \right| \tag{5}$$

After estimating the probability values, a new updating position equation is employed to update the binary position vector of each individual, as shown in Eq. 6.

$$x_i^d(t+1) = \begin{cases} complement(x_i^d(t)) & ,if\ rand \le V(x_i^d(t+1)) \\ x_i^d(t) & ,otherwise \end{cases} \tag{6}$$

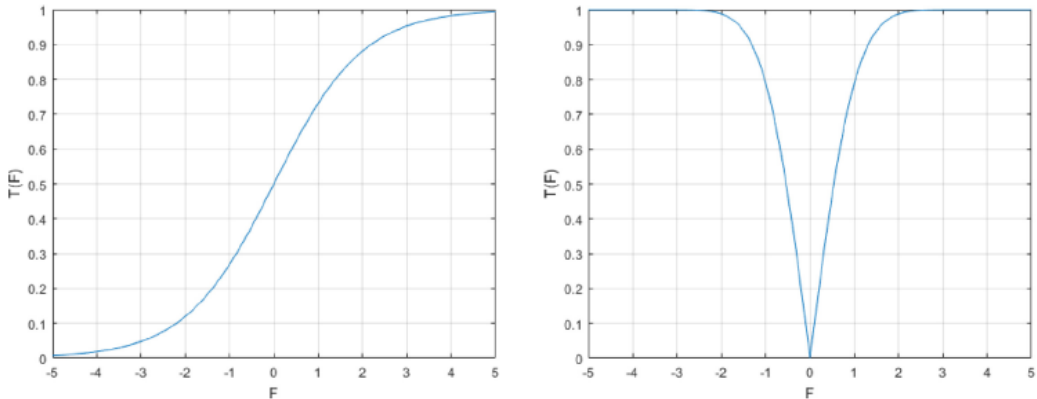

Figure 1. Figure 2. (a) S-shaped transfer function (b) V-shaped transfer function

## 5. BINARY SINE COSINE ALGORITHM FOR THE FEATURE SELECTION PROBLEM

Feature selection is a process of selecting relevant features of a dataset in order to improve the learning performance, decreasing the computational complexity, and building a better classification model. Based on the nature of the feature selection problem, a binary algorithm is usually applied to find an optimum feature subset. Every individual in the binary algorithms is represented as a binary vector with N entries, where N is the total number of features in a dataset. Each vector has the value 0 or 1, where zero indicates that the feature is not selected whereas one represents that the feature is selected. For this reason, in this work, two proposed binary versions of the SCA are applied in the feature selection problem.

Feature selection can be considered as a multi-objective problem in which two contrary objectives must be satisfied. These two objectives are the maximum accuracy, and the other is the minimum number of selected features. The fitness function that is used to evaluate each individual is shown in Eq.7.

$$Fitness = \alpha E_R(D) + \beta \frac{|R|}{|C|} \qquad (7)$$

where $E_R(D)$ is the classification error, $|R|$ is the number of selected features, $|C|$ is the total number of features in the dataset, $\alpha$ and $\beta$ are two parameters related to the importance of accuracy and number of selected features, $\alpha \in [0, 1]$ and $\beta=1-\alpha$ [29].

# 6. EXPERIMENTAL EVALUATION

## 6.1. Experimental settings

In this section, the performance of the SBSCA and VBSCA algorithms are evaluated and compared to other binary algorithm exists in the literature. In order to validate the experiment, five UCI datasets are selected with various number of features and instances. Table 1 depicts the details of each dataset. For the evaluation process, each dataset is split into %80 for training and %20 for testing. All the experiments were repeated for 30 runs to obtain meaningful results. In this work, k-nearest neighbor classifier (KNN) is used to indicate the classification error rate of the selected feature subset with k=5. All the experiments are performed on PC with Intel Core(TM) i7-3770 3.4GHz CPU and 8.00 GB RAM using MATLAB 2014 software.

The proposed SBSCA and VBSCA are compared with BBA [42], BGSA, BGWO, and BDA. The initial and specific parameters of each algorithm are reported in Table 2.

Table 1. List of datasets used in the experiment

| Dataset | No. of features | No. of instances | No. of classes |
|---|---|---|---|
| Pima | 9 | 768 | 2 |
| Breast Cancer | 10 | 683 | 2 |
| Heart | 14 | 270 | 2 |
| Lymphography | 19 | 148 | 4 |
| Breast-WDBC | 31 | 569 | 2 |

In order to have a fair comparison, all the algorithms use the same initial settings. Each algorithm is randomly initialized with the population size, and the number of iterations are set to 20 and 300. The $\alpha$ parameter in the fitness function has a value of 0.99. Evaluation criteria for all the algorithms are considered as average classification accuracy and number of selected features. To prove the significance of the proposed algorithms over other algorithms a non-parametric statistical test called Friedman test [43] is conducted as well.

Table 2. Parameter settings for algorithms

| Algorithm | Parameter | Value |
|---|---|---|
| BBA | $Q_{min}$ | 0 |
| | $Q_{max}$ | 2 |
| | A | 0.5 |
| | r | 0.5 |
| BGSA | $G_0$ | 100 |
| bGWO | a | [2 0] |
| bSCA | a | 2 |

## 6.2. Numerical Results

In this section, the results of the proposed binary versions of the SCA, SBSCA, and VBSCA are compared with other binary metaheuristic algorithms which are widely used to solve the feature selection problem. Table 3 outlines the result of BBA, BGSA, BGWO, BDA, SBSCA, and VBSCA based on the average and standard deviation of the accuracy. Note that the best results are highlighted in bold. As per results reported in Table 3, the SBSCA algorithm provides the competitive or even better results on all the datasets. It achieves the same results like BGSA on Pima and Breast Cancer datasets, while outperforms all other algorithms on Heart, Lymphography, and Breast-WDBC datasets.

Table 3. Comparison between the SBSCA, VBSCA and other binary metaheuristic algorithms based on average accuracy

| Dataset | | BBA | BGSA | BGWO | BDA | SBSCA | VBSCA |
|---|---|---|---|---|---|---|---|
| Pima | AVE | 0.7541 | **0.7727** | 0.7667 | 0.6697 | **0.7727** | **0.7727** |
| | STD | 0.0119 | 0.0000 | 0.0098 | 0.1120 | 0.0000 | 0.0000 |
| Breast Cancer | AVE | 0.9983 | **1.0000** | 0.9998 | 0.8659 | **1.0000** | **1.0000** |
| | STD | 0.0031 | 0.0000 | 0.0013 | 0.1038 | 0.0000 | 0.0000 |
| Heart | AVE | 0.8525 | 0.8772 | 0.8716 | 0.6975 | **0.8963** | 0.8926 |
| | STD | 0.0179 | 0.0091 | 0.0257 | 0.2372 | 0.0092 | 0.0075 |
| Lymphography | AVE | 0.7978 | 0.8344 | 0.8300 | 0.7978 | **0.8767** | 0.8633 |
| | STD | 0.0230 | 0.0205 | 0.0268 | 0.2174 | 0.0250 | 0.0202 |
| Breast-WDBC | AVE | 0.9518 | 0.9591 | 0.9532 | 0.9556 | **0.9673** | 0.9655 |
| | STD | 0.0060 | 0.0048 | 0.0066 | 0.0679 | 0.0046 | 0.0022 |
| Friedman Test | | 5.30 | 2.60 | 4.20 | 5.50 | **1.40** | 2.00 |

Table 4 shows the average and standard deviation of the number of selected features. It can be observed that the two proposed binary algorithms nearly have the same performance in term of the number of selected features. Moreover, the SBSCA, VBSCA, and the BDA obtained the best result on the Breast cancer dataset, while BDA had a competitive result on the Breast-WDBC and Lymphography datasets.

Table 4. Comparison between the SBSCA, VBSCA and other binary metaheuristic algorithms based on average number of selected features

| Dataset | | BBA | BGSA | BGWO | BDA | SBSCA | VBSCA |
|---|---|---|---|---|---|---|---|
| Pima | AVE | **3.00** | 5.00 | 5.10 | 5.00 | 5.00 | 5.00 |
| | STD | 1.53 | 0.00 | 0.31 | 0.00 | 0.00 | 0.00 |
| Breast Cancer | AVE | 3.27 | 3.20 | 4.27 | **3.00** | **3.00** | **3.00** |
| | STD | 1.41 | 0.41 | 1.08 | 0.00 | 0.00 | 0.00 |
| Heart | AVE | 5.07 | **4.97** | 7.03 | 5.33 | 5.27 | 5.27 |
| | STD | 2.07 | 1.16 | 0.76 | 1.42 | 1.46 | 1.41 |
| Lymphography | AVE | 6.33 | 7.63 | 9.73 | **6.03** | 6.13 | 7.23 |
| | STD | 3.07 | 1.73 | 2.21 | 1.33 | 1.55 | 2.06 |
| Breast-WDBC | AVE | 11.10 | 12.77 | 11.93 | 4.27 | **4.20** | 9.33 |
| | STD | 3.39 | 2.51 | 2.46 | 1.14 | 1.06 | 2.25 |
| Friedman Test | | 3.00 | 3.90 | 5.80 | 2.70 | **2.40** | 3.20 |

# 7. CONCLUSIONS

In this paper, two binary variants of the Sine Cosine Algorithm (SCA) were proposed and used to find the effective features in the wrapper approach. The continuous version of the SCA is converted either by S-shaped and V-shaped transfer functions to develop two binary algorithms SBSCA and VBSCA. The proposed algorithms are employed in feature selection problem for disease detection using KNN classifier. The SBSCA, VBSCA and the four state-of-the art binary algorithms are applied on five medical datasets from UCI repository and their results are compared. The experimental results show that the SBSCA algorithm is able to compete and/or achieves better results compared to other algorithms on most of the datasets. For future studies, the bSCA versions can be applied to various public datasets with different classifiers, and real-world problems. Furthermore, it would be interesting to use the bSCA in solving problems with multiple objectives.

## AUTHORS

Dr. Mohammad-Hossein Nadimi-Shahraki received his Ph.D. in computer science-artificial intelligence from University Putra of Malaysia (UPM) in 2010. Currently, he is an Ass professor in faculty of computer engineering and a senior data scientist in Big Data Research Center in Islamic Azad University of Najafabad (IAUN). His research interests include big data analytics, data mining algorithms, medical data mining, machine learning and metaheuristic algorithms.

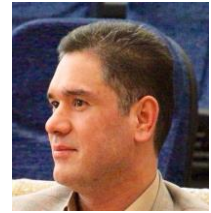

Ms. Shokooh Taghian was born in Iran. She received M.S. degrees in computer software engineering from the faculty of computer engineering in IAUN. She is currently researching as a research assistant and developer in Big Data Research Center in (IAUN). Her research interests focus on metaheuristic algorithms, machine learning and medical data analysis.

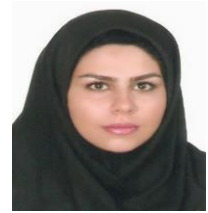